# UrBLiMP: A Benchmark for Evaluating the Linguistic Competence of Large Language Models in Urdu

**Farah Adeeba**[1,2]    **Brian Dillon**[1]    **Hassan Sajjad**[3]    **Rajesh Bhatt**[1]

[1]Department of Linguistics, University of Massachusetts Amherst, USA
[2]Department of Computer Science, University of Engineering and Technology, Lahore, Pakistan
[3]Department of Computer Science, Dalhousie University, Canada
`fadeeba@umass.edu, bwdillon@umass.edu, hsajjad@dal.ca, bhatt@umass.edu`

## Abstract

Multilingual Large Language Models (LLMs) have shown remarkable performance across various languages; however, they often include significantly less data for low-resource languages such as Urdu compared to high-resource languages like English. To assess the linguistic knowledge of LLMs in Urdu, we present the Urdu Benchmark of Linguistic Minimal Pairs (UrBLiMP) i.e. pairs of minimally different sentences that contrast in grammatical acceptability. UrBLiMP comprises 5,696 minimal pairs targeting ten core syntactic phenomena, carefully curated using the Urdu Treebank and diverse Urdu text corpora. A human evaluation of UrBLiMP annotations yielded a 96.10% inter-annotator agreement, confirming the reliability of the dataset. We evaluate twenty multilingual LLMs on UrBLiMP, revealing significant variation in performance across linguistic phenomena. While LLaMA-3-70B achieves the highest average accuracy (94.73%), its performance is statistically comparable to other top models such as Gemma-3-27B-PT. These findings highlight both the potential and the limitations of current multilingual LLMs in capturing fine-grained syntactic knowledge in low-resource languages.

## 1 Introduction

Large Language Models (LLMs) have become crucial components of Natural Language Processing (NLP) systems, enabling a wide variety of tasks including summarization, machine translation, and dialogue generation. While evaluation efforts have largely focused on LLMs' performance in tasks requiring world knowledge and general language understanding, the extent to which these models acquire specific linguistic phenomena remains insufficiently explored—particularly for many low-resource languages.

For the evaluation of linguistic knowledge of LMs, a prominent and effective methodology involves the use of minimal pairs(Warstadt et al., 2020) : sequences of words that differ minimally, with only one forming an acceptable sentence (e.g., *The cat sleeps* vs. *The cat sleep*). LMs are evaluated on a pairwise zero-shot forced choice task, measuring whether they assign a higher probability to the acceptable versus the unacceptable sequence.

The Benchmark of Linguistic Minimal Pairs (BLiMP) (Warstadt et al., 2020), for instance, comprises 67,000 such pairs for English, semi-automatically generated to evaluate a wide range of linguistic phenomena and assess LMs' knowledge of grammar on a large scale. Similar resources have been developed for other high-resource languages, including CLiMP (Xiang et al., 2021) and SLING (Song et al., 2022) for Chinese, TurBLiMP (Başar et al., 2025) for Turkish, RuBLiMP (Taktasheva et al., 2024) for Russian, JBLiMP (Someya and Oseki, 2023) for Japanese, and BLiMP-NL for Dutch.

Despite the increasing success of multilingual LLMs, the question of which linguistic phenomena they can or cannot learn remains poorly understood for many languages, especially those with limited digital resources. For Urdu, comparable diagnostic resources for fine-grained linguistic evaluation are critically scarce in both coverage and granularity. While some benchmarking efforts exist (Tahir et al., 2025), they largely focus on functional competence. Even a dataset targeting ergativity in Hindi (Kryvosheieva and Levy, 2024) has limited direct applicability to Urdu due to distinct script and vocabulary. Furthermore, MultiBLiMP (Jumelet et al., 2025), a multilingual extension of BLiMP, includes a small Urdu component, but with only approximately 1,000 examples focusing exclusively on subject agreement, its scope is insufficient to capture the intricate syntactic richness of the language. A comparative summary of these

| | Language | Size | P | Method |
|---|---|---|---|---|
| BLiMP | English | 67K | 67 | Dict & Templates |
| CLiMP | Chinese | 16K | 16 | Translation & Templates |
| SLING | Chinese | 38K | 38 | UD & Templates |
| TurBLiMP | Turkish | 16K | 16 | Semi-automatically |
| RuBLiMP | Russian | 45K | 45 | Semi-automatically |
| JBLIMP | Japanese | 331 | 39 | Extracted from articles |
| BLiMP-NL | Dutch | 8.4K | 84 | Semi-automatically |
| MultiBLiMP | 101 Languages | 128K | 2 | UD |
| BHS | Basque, Hindi, Swahili | 300 | 3 | Dictionary & Templates |
| **UrBLiMP** | **Urdu** | **5696** | **19** | **Text corpus, rules** |

Table 1: Comparison of existing minimal pair datasets across languages. P is number of paradigms, UD: Universal Dependencies, BHS: work by (Kryvosheieva and Levy, 2024)

datasets is presented in Table 1.

This work aims to bridge this significant gap in Urdu by introducing **UrBLiMP**—the *Urdu Benchmark of Linguistic Minimal Pairs*—a syntactically informed diagnostic dataset for evaluating LLMs' grammatical competence in Urdu. UrBLiMP comprises 5,696 minimal pairs spanning ten core syntactic phenomena. These were constructed through a hybrid methodology combining treebank-derived templates with surface pattern matching applied to diverse raw corpora, followed by rigorous manual verification.

By providing this benchmark, we move beyond generic accuracy metrics and offer a linguistically grounded tool for identifying systematic weaknesses in LLMs' understanding of Urdu grammar. UrBLiMP thus lays the foundation for advancing both the evaluation and improvement of multilingual models in the context of low-resource, morphologically rich languages like Urdu.

## 2 UrBLiMP

The process of dataset creation is outlined in this section. The selected linguistic phenomena and corresponding paradigms are shown in Table 2. Minimal pairs were constructed using both the Urdu Treebank and surface-level patterns applied to raw text. For each phenomenon, transformation rules were formulated to produce minimal pairs differing on a single grammatical property. The following subsections describe the selected linguistic phenomenon, minimal pairs generation and transformation procedures in more detail.

### 2.1 Linguistic Phenomenon

UrBLiMP covers ten major linguistic phenomena of Urdu, as summarized in Table 2. Several paradigms are designed around a linguistic phenomenon to check the LMs robustness.

**Aspect Agreement** phenomena capture the restrictions governing co-occurrence of aspectual markers in Urdu. Specifically, while the habitual participle *-ta:* (تا-/تی-/تے-) may combine with the progressive marker *raha:* (رہا، رہی، رہے) to express ongoing habitual actions, its combination with the perfective aspect verb *cuka:* (چکا) is ungrammatical. The mismatch arises because the habitual aspect denotes iterative or unbounded events, whereas the perfective aspect verb implies a bounded and completed action (Butt and Rizvi, 2008), leading to a semantic conflict.

**Dative Object** phenomena investigate the ungrammaticality that results from omitting the dative postposition *ko* (کو) when it is required to mark direct objects in Urdu. Two paradigms were considered: 1) the direct object is a proper noun and 2) the direct object is a pronoun.

**Ergativity** in Urdu is a split system primarily governed by aspect. In perfective transitive constructions, the subject takes ergative marking *ne* (نے) and the verb agrees with the direct object. In non-perfective aspects (e.g., habitual or progressive), the subject is nominative and the verb agrees with it. To assess model sensitivity to this system, minimal pairs across three paradigms are created:

1. *Aspect Sensitivity:* Ungrammaticality was introduced by replacing the perfective verb with a habitual verb while retaining the ergative subject. This results in a mismatch between aspect and subject marking.

2. *Verb-Object Agreement:* In perfective transitive clauses, the verb should agree in gender with the direct object rather than the ergative-marked subject. Ungrammatical variants were created by forcing verb agreement with the subject instead.

3. *Differential Object Marking (DOM) in Ergative Constructions:* In Urdu ergative constructions (i.e., perfective transitive clauses with the subject marked by *ne*), when the direct object is marked with *ko*, the verb typically defaults to masculine singular form, regardless of the object's gender or number. To test sensitivity to this default agreement rule, ungrammatical sentence variants were constructed by replacing the expected masculine singular verb with a feminine form that agrees with the feminine object.

| Phenomenon | N | Grammatical Sentence | Ungrammatical Sentence |
|---|---|---|---|
| Aspect | 1 | ہاں وہ گورنر سے ملتا رہا تھا۔<br>hā voh gavarnar se milta: raha: tha:<br>Yes he governor from meet-IMPF PROG.PART was<br>'Yes, he had been meeting the governor.' | ہاں وہ گورنر سے ملتا چکا تھا۔<br>hā voh gavarnar se milta: cuka: tha:<br>Yes he governor from meet-IMPF PERF.PART was<br>'Yes, he had already met the governor.' |
| Dative Object | 2 | پہرےدار نے تمام ماجرا راجا کو سنایا۔<br>pahreda:r ne tama:m ma:jra: ra:ja: ko suna:ya:<br>guard ERG whole incident Raja DAT narrated.PST<br>'The guard narrated the whole incident to Raja.' | پہرےدار نے تمام ماجرا راجا سنایا۔<br>pahreda:r ne tama:m ma:jra: ra:ja: suna:ya:<br>guard ERG whole incident Raja narrated.PST<br>'The guard narrated the whole incident to Raja.' |
| Ergativity | 3 | پہرےدار نے تمام ماجرا راجا کو سنایا<br>pahreda:r ne tama:m ma:jra: ra:ja: ko suna:ya:<br>guard ERG whole incident Raja DAT narrated.PST<br>'The guard narrated the whole incident to Raja.' | پہرےدار نے تمام ماجرا راجا کو سناتا<br>pahreda:r ne tama:m ma:jra: ra:ja: ko suna:ta:<br>guard ERG whole incident Raja DAT narrates.PRES<br>'The guard narrated the whole incident to Raja.' |
| Experiencer Subject | 1 | مجھے یہ بلاگ پوسٹ پسند آیا۔<br>mujhe yeh blog post pasand a:ya:<br>to.me this blog post liking came<br>I liked this blog post. | میں یہ بلاگ پوسٹ پسند آیا۔<br>maĩ yeh blog post pasand a:ya:<br>I this blog post liking came<br>I liked this blog post. |
| Honorific | 1 | بہن جی مجھ سے ناراض تھیں<br>behan ji mujh se nara:z thĩ<br>sister from.me upset was (hon.)<br>Sister was upset with me. | بہن جی مجھ سے ناراض تھی<br>behan ji mujh se nara:z thi<br>sister from.me upset was<br>Sister was upset with me. |
| N-J Agr | 1 | ہر مجسمہ بیس فٹ اونچا ہے۔<br>har mujasma bi:s fit o:nca: hai<br>every statue twenty feet tall is<br>Every statue is twenty feet tall. | ہر مجسمہ بیس فٹ اونچی ہے۔<br>har mujasma bi:s fit onci: hai<br>every statue twenty feet tall.F is<br>Every statue is twenty feet tall. |
| Participial Relatives | 1 | آسانی سے کھولا گیا دروازہ بند کیا جا سکتا ہے<br>a:sani: se khola: gaya: darwaza: band kiya: ja: sakta: hai<br>easy-ADV from open-PTCP go-PTCP.M.SG door close do-PERF.M.SG go can be<br>'The door that was easily opened can be closed.' | آسانی سے کھولتا گیا دروازہ بند کیا جا سکتا ہے<br>a:sani: se kholta: gaya: darwaza: band kiya: ja: sakta: hai<br>easy-ADV from open-IPFV.M.SG go-PTCP.M.SG door close do-PERF.M.SG go can be<br>'The door that kept opening easily can be closed.' |
| Subj-Verb Agr | 3 | اور اس کا باپ تو شاید پاگل ہو جاتا۔<br>aur us ka ba:p to shayad pagal ho jata:<br>and 3SG.M GEN.M father FOCUS perhaps mad become COND.M.SG<br>'And his father might have gone mad.' | اور اس کا باپ تو شاید پاگل ہو جاتی۔<br>aur us ka ba:p to shayad pagal ho jati:<br>and 3SG.M GEN.M father FOCUS perhaps mad become COND.F.SG<br>'And his father (incorrectly) might have gone mad (feminine verb).' |
| Order Variation | 1 | جو بات ہے وہ بولو۔<br>jo ba:t hai vo bolo<br>what matter is that say.IMP.PL<br>'Say what the matter is.' | جو ہے وہ بولو بات۔<br>jo hai vo bolo ba:t<br>what is that say matter<br>'Say what is that matter.' (incorrect order) |

Table 2: Ten Urdu linguistic phenomena covered in this study, with grammatical and ungrammatical sentence examples. Minimal contrasts are emphasized. The second line of each example shows transliterated Urdu, and the third line provides an English translation. *N* indicates the number of paradigms within each phenomenon.

**Obliqueness** phenomenon tests models for case-driven contrasts where oblique forms are required due to syntactic context. We construct ungrammatical variants by replacing expected oblique forms with direct (non-oblique) forms across various categories, including masculine singular nouns, plural nouns, adjectives, verbs, and pronouns.

**Experiencer Subjects.** In Urdu, experiencer-subject constructions typically involve a subject marked in the dative case, often with the postposition *ko*. However, when the subject is a pronoun, the dative case is realized through dedicated oblique pronominal forms (e.g., *mujhe*, *tujhe*), and the postposition *ko* is not overtly used.

In this paradigm, we focus on constructions where the experiencer is a pronominal subject in its dative (oblique) form . Ungrammatical variants were generated by replacing these forms with nominative pronouns (e.g., *meN*) or incorrect oblique forms that lack proper dative alignment (e.g., *us* without *ko*).

**Subject-Verb Agreement** phenomenon is tested across three paradigms: number, gender, and person agreement. Ungrammatical sentence variants were generated by violating the expected agreement between the subject and the verb in each paradigm.

**Honorific.** Urdu marks respect or politeness by enforcing plural agreement on the verb when addressing or referring to someone honorifically. This applies even when the referent is singular, especially in constructions involving titles or respectful nouns such as (ji:, muhtram, muhtarma:, ja:n)محترم, محترمہ, جی, or جان. To evaluate this phenomenon, ungrammatical variants were created by replacing the required plural verb forms with singular forms, thereby violating the honorific agreement pattern.

**Participial Relatives** phenomenon involves the use of perfective participles in relative clauses that modify a noun e.g. *khola gaya darvaza:(opened door)*. In Urdu, such constructions typically require a passive participial verb form to ensure grammaticality. To test this, ungrammatical variants were generated by replacing the participial verb with an imperfective or active form, disrupting the required agreement and aspectual constraints within the relative clause. **Word Order Variation** Urdu generally allows relatively flexible word order; however, certain positions are preferred for clarity and naturalness, especially in relative clauses. In this paradigm, ungrammatical or less acceptable variants were created by altering the canonical word order, such as shifting the noun out of a relative clause to the clause-final position, which disrupts the natural syntactic structure and interpretability of the sentence.

## 2.2 Minimal Pairs Generation

Minimal pairs were generated using naturally occurring sentences extracted from Urdu texts. Two primary resources were utilized for this purpose: the Urdu Treebank and in-house Urdu Corpus.

### 2.2.1 Urdu Treebank

To extract sentences exhibiting the targeted linguistic phenomena, the Urdu Treebank (Ehsan and Hussain, 2021) was utilized. This resource comprises 7,854 Urdu sentences annotated in the Penn Treebank style. A subset containing relevant syntactic structures was selected using pattern-based search.

For example, to extract sentences illustrating the Aspect Agreement phenomenon, the syntactic pattern *(VC (VBF) (AUXP))* was used. From the matching sentences, those containing a VBF verb with a habitual participle suffix (e.g., -ta) were retained. These extracted sentences were then manipulated to construct minimal pairs: the auxiliary phrase (AUXP) — such as *raha:, rahi:* — was replaced with *cuka: or cuki:* to generate ungrammatical variants.

### 2.2.2 Urdu Corpus

Since a sufficient number of required sentences could not be obtained from the Urdu Treebank, additional data were extracted from an in-house collected corpus of Urdu text. This corpus comprises approximately 735 million tokens and includes diverse web-based content from sources such as Common Crawl, Twitter, and news articles.

For each paradigm, specific regular expressions were crafted to extract potentially relevant sentences. For example, to retrieve candidate sentences for the word order variation phenomenon, the regular expression ^جہ\b.*\b وہ \b was used. The extracted sentences were then manually reviewed, and only those that conformed to the targeted linguistic paradigm were retained. Furthermore, data cleaning was performed to correct typographical errors and remove unnecessary space insertions or deletions. Finally, minimal pairs were constructed from the selected sentences by systematically introducing controlled variations.

## 2.3 Human Evaluation

To assess the quality of the minimal pairs generated, two rounds of human validation were performed using the PCIbex platform (Zehr and Schwarz, 2018). A total of 17 native Urdu speakers participated in the evaluation. The group included six males and eleven females, with ages ranging from 23 to 46 years. Among the participants, four were linguists, while the remaining evaluators had at least a high school education, including three with Ph.D.-level qualifications.

In the first round, all annotators underwent a brief training phase in which they at least annotated 20 pairs of demonstration sentences. This familiarization step helped ensure a consistent understanding of the evaluation interface and task.

In the second round, each annotator labeled approximately 190 pairs, covering about 10 pairs from each linguistic paradigm. The sentences of different paradigms were randomly shuffled. To ensure the reliability of judgments, each sentence pair was annotated by at least three different evaluators. The complete annotation task took approximately one hour per annotator.

The final raw human accuracy mean overall paradigms is 96.10% . The inter-annotator agreement as measured by Fleiss' kappa is 0.89, indicating almost perfect agreement according to (Landis and Koch, 1977). The average acceptability accuracy per paradigm is shown in Table 9.

## 3 Experimental Setup

This section presents the experimental setup used to evaluate the syntactic understanding of various multilingual and Urdu language models. We detail the selected models and describe the accuracy-based evaluation metric applied across linguistic phenomena.

### 3.1 Evaluation Models

For evaluation, we employ a diverse set of 20 multilingual language models as listed in Table 3. These include encoder-only models such as multilingual **BERT** (Devlin et al., 2018), encoder–decoder models like **mT5** (Xue et al., 2021), and decoder-only models including **LLaMA3** (Grattafiori et al., 2024) and **Gemma** (Team, 2025). The Gemma models are particularly noteworthy for being trained on a balanced multilingual corpus encompassing 140 languages.

We also consider **BLOOMZ** (Muennighoff et al., 2022), **DeepSeek** (DeepSeek-AI, 2025) and the **Granite** series (Granite Team, 2024) models. Moreover, a notable addition is **Alif-1.0-8B-Instruct** (Traversaal, 2025), a publicly available, continually pre-trained Urdu model based on `unsloth/Meta-Llama-3.1-8B`.

### 3.2 Evaluation Measure

Following prior work (Song et al., 2022; Jumelet et al., 2025), we perform evaluation at the sentence level, rather than at the position of the key linguistic items, which was the focus of earlier studies. Consistent with (Song et al., 2022), we compute perplexity(Holtzman et al., 2021) for causal language models and pseudo-perplexity for masked language models over each linguistic pair.

| Model | Size | IT | Source |
|---|---|---|---|
| gemma-3-1b-it | 1B | Yes | (Team, 2025) |
| gemma-3-4b-it | 4B | Yes | (same as above) |
| gemma-3-12b-it | 12B | Yes | (same as above) |
| gemma-3-27b-it | 27B | Yes | (same as above) |
| gemma-3-1b-pt | 1B | No | (same as above) |
| gemma-3-4b-pt | 4B | No | (same as above) |
| gemma-3-12b-pt | 12B | No | (same as above) |
| gemma-3-27b-pt | 27B | No | (same as above) |
| Llama-3-8B | 8B | No | (Grattafiori et al., 2024) |
| Llama-3-70B | 70B | No | (same as above) |
| DS | 70B | No | (DeepSeek-AI, 2025) |
| Alif-1.0-8B-Instruct | 8B | Yes | (Traversaal, 2025) |
| granite-3.3-2b-base | 2B | No | (Granite Team, 2024) |
| granite-3.3-8b-base | 8B | No | (same as above) |
| granite-3.3-8b-instruct | 8B | Yes | (same as above) |
| granite-3.3-2b-instruct | 2B | Yes | (same as above) |
| bloomz-7b1-p3 | 7.1B | No | (Muennighoff et al., 2022) |
| BERT | 110M | No | (Devlin et al., 2018) |
| mt5-small | 300M | No | (Xue et al., 2021) |
| mt5-large | 1.23B | No | (same as above) |

Table 3: The set of multilingual language models evaluated in this work. IT = Instruction Tuned. BERT=bert-base-multilingual-cased DS=DeepSeek-R1-Distill-Llama-70B

To evaluate a language model, we use **accuracy** as the primary metric. Accuracy is defined as the proportion of minimal pairs in which the model assigns a lower (pseudo-)perplexity to the grammatically acceptable sentence.

Formally, let $N$ be the total number of sentence pairs $(s_{\text{good}}, s_{\text{bad}})$, where $s_{\text{good}}$ is the acceptable sentence and $s_{\text{bad}}$ is the unacceptable counterpart. Let $\text{ppl}(s)$ denote the (pseudo-)perplexity of sentence $s$. Then the accuracy is computed as:

$$\text{Accuracy} = \frac{1}{N} \sum_{i=1}^{N} \mathbb{I}\left[\text{ppl}(s_{\text{good}}^i) < \text{ppl}(s_{\text{bad}}^i)\right]$$

where $\mathbb{I}[\cdot]$ is the indicator function that returns 1 if the condition holds and 0 otherwise.

## 4 Results and Analysis

Table 4 presents the results of the language models (LMs) and human performance on each syntactic phenomenon. Overall, LLaMA-3-70B achieves the highest average performance across most paradigms. However, there is no single LM that consistently outperforms all others across all

| | Aspect Agr | Dative-Obj | Ergativity | Exper-Sub | Honorific | N-J Agr | Oblique | PR | S-Verb Agr | order varion | Average |
|---|---|---|---|---|---|---|---|---|---|---|---|
| gemma-3-1b-pt | 100.00 | 98.62 | 87.36 | 79.01 | 73.20 | 93.00 | 87.34 | 79.07 | 87.60 | 82.18 | 86.74 |
| gemma-3-4b-pt | 100.00 | 98.24 | 90.15 | 84.20 | 73.86 | 93.00 | 85.58 | 81.06 | 88.91 | 86.14 | 88.11 |
| gemma-3-12b-pt | 100.00 | 98.24 | 93.41 | 94.07 | 84.97 | 96.50 | 90.20 | 85.71 | 92.58 | 91.09 | **92.68** |
| gemma-3-27b-pt | 100.00 | **99.62** | 94.94 | 84.20 | 84.31 | **97.00** | 92.75 | 85.38 | **94.89** | 91.09 | **92.42** |
| gemma-3-1b-it | 99.00 | 89.81 | 70.02 | 62.72 | 55.56 | 76.50 | 70.07 | 57.48 | 70.44 | 63.37 | 71.50 |
| gemma-3-4b-it | 99.75 | 92.65 | 86.62 | 56.54 | 74.51 | 77.00 | 76.78 | 74.42 | 82.19 | 68.32 | 78.88 |
| gemma-3-12b-it | 100.00 | 95.16 | 91.72 | 76.05 | 79.74 | 89.50 | 79.83 | 78.74 | 89.99 | 78.22 | 85.89 |
| gemma-3-27b-it | 100.00 | 94.40 | 93.98 | 56.05 | 84.97 | 92.50 | 85.05 | 85.38 | 92.55 | 83.17 | 86.81 |
| Llama-3-8B | 100.00 | 92.60 | 91.45 | **98.27** | 93.46 | 89.50 | 85.14 | 90.03 | 76.60 | 90.10 | 90.71 |
| Llama-3-70B | 100.00 | 96.43 | **96.38** | 95.80 | 98.04 | 94.00 | **92.77** | 92.69 | 87.13 | 94.06 | **94.73** |
| Alif-1.0-8B-Instruct | 100.00 | 91.26 | 93.61 | 98.02 | 98.04 | 89.00 | 90.73 | 91.69 | 83.13 | **97.03** | 93.25 |
| granite-3.3-8b-base | 97.51 | 88.83 | 88.67 | 91.11 | 90.20 | 85.00 | 84.83 | 82.72 | 72.48 | 93.07 | 87.44 |
| granite-3.3-2b-base | 99.25 | 89.59 | 82.07 | 77.04 | 83.66 | 83.00 | 82.25 | 80.07 | 64.78 | 85.15 | 82.69 |
| granite-3.3-8b-instruct | 98.75 | 89.21 | 84.14 | 60.74 | 94.77 | 80.50 | 77.92 | 74.09 | 71.06 | 91.09 | 82.23 |
| granite-3.3-2b-instruct | 99.00 | 80.53 | 80.18 | 92.10 | 88.89 | 74.00 | 74.04 | 78.41 | 61.07 | 76.24 | 80.45 |
| DeepSeek[1] | 100.00 | 88.20 | 93.46 | 96.30 | **99.35** | 90.50 | 91.66 | 88.70 | 86.33 | 90.10 | 92.46 |
| bloomz-7b1-p3 | 99.25 | 96.28 | 90.85 | 67.65 | 84.31 | 87.50 | 84.91 | 81.06 | 84.58 | 85.15 | 86.15 |
| Bert[2] | 70.82 | 75.44 | 79.91 | 82.02 | 80.97 | 81.36 | 80.95 | 79.50 | 69.73 | 90.10 | 78.77 |
| mt5-small | 27.43 | 29.77 | 54.75 | 60.62 | 36.60 | 46.50 | 60.67 | 79.40 | 50.79 | 60.40 | 50.69 |
| mt5-large | 71.82 | 58.12 | 58.14 | 79.75 | 51.50 | 56.48 | 63.56 | 73.52 | 58.30 | 62.38 | 63.36 |
| **Human** | **100** | **94.27** | **97.71** | **93.89** | **94.17** | **97.29** | **95.46** | **95.63** | **95.68** | **98.75** | |

Table 4: Percentage accuracy of various models on different syntactic phenomena. Average accuracy accross all phenomena, it = Instruction Tuned, pt = Pretrained, PR = Participial Relatives

[1] DeepSeek-R1-Distill-Llama-70B

[2] bert-base-multilingual-cased

categories. In several instances, other LMs surpass LLaMA-3-70B on specific syntactic phenomena.

A comparison between multilingual models and the continual-pretrained model on Urdu (i.e., Alif-LLaMA-8B) reveals that Alif achieves a higher average accuracy than its base model, LLaMA-3-8B, based on its performance across the linguistic phenomena in Table 4. While Alif does not consistently outperform all models on individual phenomena, it shows a notable improvement on the Word Order Variation phenomenon and achieves comparable performance on the Experiencer-Subject construction. These findings underscore the benefits of continued pretraining on Urdu data.

### 4.1 Long Distance Agreement

In Urdu, gender agreement between the subject and verb is crucial. For example in (1) verb is separated from the noun. This indicates that the capability of the model to capture and utilize long-range syntactic relationships is limited, which negatively impacts its overall performance on such linguistic phenomena.

We observed that the language models struggle particularly with long-distance agreement. Specifically, in cases such as gender paradigm in phenomenon of subject-verb agreement, when the distance between the subject and the verb increases, the ability of the model to correctly predict gender agreement significantly deteriorates.

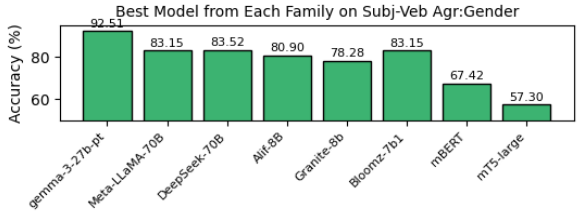

Figure 1: Best-performing model from each architecture family on the subject-Verb Agreement in gender paradigm.

For example (2) is marked as ungrammatical by most of the LMs even with the LlaMA-70B model because there is long distance between subject استانی (teacher.F) and verb سوتی(sleep.F).

(1) فیصل اپنے سکول سے خوشی خوشی گھر واپس آیا
Faisal apne sku:l sy Kushi Kushi gher aya:
Faisal.M his.POSS school from happily home came-Perf.M

Faisal came home happily from his school

(2) استانی سبق پڑھا کر آرام سے کرسی پر بیٹھ کر سوتی۔
Ustani: sabq peRha: kar a:ra:m se kursi: par beTh kar soti:
teacher lesson teach-Perf.F do-Perf.F comfortably chair on sit-Perf.F sleep-Perf.F

'After teaching the lesson, the teacher would sit comfortably on a chair and sleep.'

Figure 1 presents a comparison of the best-performing model from each architecture family evaluated on the gender paradigm of Subj-Verb phenomenon . The `gemma-3-27b-pt` model achieved the highest accuracy , substantially outperforming others, including LLaMA-3-70B , DeepSeek's LLaMA-70B , `Alif-8B-Instruct` and `Bloomz-7b1-p3`.

In contrast, smaller encoder-based models such as `bert-base-multilingual-cased` and `mt5-large` achieved considerably lower accuracy (67.42% and 57.30%, respectively), indicating the limitations of encoder-only or smaller multilingual models for fine-grained gender-related linguistic understanding.

### 4.2 Phenomenon Specific Results

A linguistic breakdown of the evaluation results reveals varying degrees of model competence across different grammatical phenomena in Urdu. Some constructions—particularly those that are categorically distinct in surface form—proved relatively straightforward for both humans and language models, whereas others remained more challenging, especially for smaller models.

Aspect Agreement emerged as one of the simplest phenomena for both human annotators and LLMs. Even smaller-scale models achieved near-perfect accuracy on this task. A likely explanation is that in Urdu, habitual and perfective aspects are rarely co-expressed within the same clause, resulting in clear distributional patterns. This distinct separation made the classification task more deterministic and less ambiguous for models.

In the case of Dative Object marking, both noun and pronoun contexts yielded accuracies above 92% across most models (see Table 7). However, models such as those in the Granite series displayed slightly reduced performance for dative nouns, achieving accuracies around 88%. Pronouns, on the other hand, were processed with higher precision, likely due to their more regular and less variable surface forms.

With respect to Ergativity, models generally succeeded in detecting aspect-sensitive agreement patterns, particularly the requirement for perfective verb forms in ergative constructions. Nonetheless, performance dropped considerably in related phenomena such as Object-Verb Agreement and Differential Object Marking (DOM). In both these cases, most models exhibited accuracy scores below 90%, suggesting that these constructions involve more nuanced syntactic and semantic dependencies, which continue to challenge current LLMs.

In Obliqueness, models struggled particularly with adjective and masculine singular nouns even its very local. The oblique case marking in Urdu—especially when realized through subtle morphological alternations—proved difficult for several models to capture consistently. Only LlaMA-70B and DeepSeek-70B accuracy is above 90%. This indicates an ongoing limitation in how models process inflectional morphology in morphologically rich languages like Urdu.

### 4.3 Model Size

We observed that model performance generally improved with increasing parameter size within the Gemma series. Notable accuracy gains were achieved between the 1B and 27B parameter models, while there is minimal decline in accuracy from 12B to 27B. The statistical significance of is gain was confirmed through a two-tailed Wilcoxon signed-rank test. Table 5 presents the results among the `gemma-PT` models to assess whether the differences in performance are statistically significant. Table 5 shows the difference between `gemma-3-12b-pt` and `gemma-3-27b-pt` is not statistically significant, suggesting that most of the performance gain saturates around the 12B model.

| LM 1 | LM 2 | p (raw) | p (Holm) | Sig |
|---|---|---|---|---|
| gemma-3-1b-pt | 4b-pt | 0.078 | 0.156 | No |
| gemma-3-1b-pt | 12b-pt | 0.007 | 0.0390 | Yes |
| gemma-3-1b-pt | 27b-pt | 0.003 | 0.0234 | Yes |
| gemma-3-4b-pt | 12b-pt | 0.007 | 0.0390 | Yes |
| gemma-3-4b-pt | 27b-pt | 0.007 | 0.0390 | Yes |
| gemma-3-12b-pt | 27b-pt | 0.460 | 0.460 | No |

Table 5: Pairwise Wilcoxon signed-rank test results for Gemma-PT models on linguistic phenomena. Holm-corrected p-values are reported.

| LM1 | LM2 | p-raw | p-holm | Sig-holm |
|---|---|---|---|---|
| Gemma-12b-pt | Gemma-27b-pt | 0.460 | 1 | No |
| Gemma-12b-pt | Llama-70B | 0.238 | 1 | No |
| Gemma-12b-pt | Alif | 0.910 | 1 | No |
| Gemma-27b-pt | Llama-70B | 0.496 | 1 | No |
| Gemma-27b-pt | Alif | 0.910 | 1 | No |
| Llama-70B | Alif | 0.195 | 1 | No |

Table 6: Pairwise comparisons of top-performing models using Wilcoxon signed-rank test with Holm correction. No statistically significant differences were found among the top models at $\alpha = 0.05$.

However, when multiple models were compared, it was found that larger size did not always correspond to better syntactic accuracy. Average accuracy of Gemma-3-12b-pt, Gemma-3-27b-pt, LLaMA-3-70B, and Alif-1.0-8B-Instruct is 92.68, 92.42, 94.73 and 93.25, respectively is comparable. As illustrated in Figure 2, the smaller model (e.g., *gemma-3-27b-pt*) was observed to outperform larger models (*Llama-3-70B and DeepSeek-R1-Distill-Llama-70B*) on several syntactic paradigms, dative object ProperNoun, dative object pronoun, noun adjective agreement and all three paradigms of subject-verb agreement. These results suggest that certain syntactic patterns were generalized more effectively by smaller models, potentially due to simpler architectures or reduced overfitting to training data idiosyncrasies. The statistical significance of this difference was confirmed through a two-tailed Wilcoxon signed-rank test. As shown in Table 6, no statistically significant differences ($\alpha = 0.05$) were observed among gemma-3-12b-pt, gemma-3-27b-pt, Meta-Llama-3-70B, and Alif, indicating that these models perform comparably in our evaluation.

### 4.4 Pretrained vs Instruction-Tuned Models

A comparative evaluation was performed on the pretrained (-pt) and instruction-tuned (-it) variants of the Gemma models across four sizes: 1B, 4B, 12B, and 27B. To evaluate the statistical significance of differences between the paired model variants, a two-tailed Wilcoxon signed-rank test was applied to the accuracy scores.

We found that the pretrained variants of Gemma-3-1B, 3-4B, and 3-12B significantly outperformed their instruction-tuned counterparts. These results suggest that instruction tuning may adversely affect syntactic generalization in smaller-scale models. Fine-grained results of pretrained and instruction-tuned models against each paradigms are shown in Appendix Table 7 and 8

## 5 Conclusion

In this study, we introduced a new benchmark to evaluate the syntactic capabilities of multilingual language models, with a focus on under-represented and morphologically rich constructions in Urdu. The dataset was constructed following the SLING framework, using naturally formulated minimal pairs that target core syntactic phenomena. It was found that generally multilingual models performed reasonably in capturing local dependencies and agreement structures. However, reduced performance was observed on complex syntactic constructions, such as long distance agreement in subject-verb. These findings highlight the challenges that remain for language modeling in low-resource and typologically diverse languages, emphasizing the need for more linguistically informed pretraining and evaluation approaches.

### Limitations

In this study, the evaluation of linguistic phenomena in Urdu was limited primarily by the availability of annotated data. The existing Urdu Treebank contains only a small number of relevant examples for the targeted syntactic categories. To address this limitation, additional instances were extracted from unannotated Urdu corpus using carefully designed regular expressions, followed by extensive manual cleaning. Despite these efforts, the dataset size remains relatively small, especially when compared to large-scale resources like the English BLiMP or the Mandarin SLING datasets. However, we believe that our proposed dataset size is sufficient to reliably evaluate the linguistic competence of language models as shown in our results.

Furthermore, the current evaluation only covers ten linguistic categories, few more phenomenon can also be included: **Reciprocity**: For instance, *bace ek dosre ko ma:r rehe h e* ( *Children are hitting each other*). This construction expresses reciprocal action between subjects and acceptable in Urdu. Its variation *bace ek dosre ma:r rehe h e* is unacceptable.

Question constructions involving relative clauses is also not included in this dataset. For instance, *jo muhabat nah e rakhta vo Xuda ko nah e janta:* ( *He who does not have love does not know God*) is acceptable but *jo kiya: nah e rakhta vo Xuda ko nah e janta:* this type of sentence includes a *wh*-word (جو) that functions as both a relative clause marker and a question-like structure, which remains unexplored in our current dataset.

These phenomena are linguistically rich and crucial for comprehensive evaluation, and their absence points to a limitation of the current resource. With improved Urdu parsers and extended annotated corpora, such constructions could be systematically included in future datasets.

## A  Ethical Consideration

We plan to open-source our dataset along with a detailed data card. The documentation will follow the templates used in the BLiMP benchmark (Warstadt et al., 2020) and the HuggingFace Datasets repository (Lhoest et al., 2021). The dataset and accompanying code will be released via a public GitHub repository under a permissive open-source license.

## B  Computational Cost

The computational cost of evaluating a language model (LM) on UrBLiMP varies depending on the model's architecture and size. Distributed inference libraries (e.g., accelerate, torchrun) can be used to optimize the process.

For a single NVIDIA A100 GPU with 80GB memory, the complete evaluation takes approximately 1.5 hours for decoder-only models and 10 hours for encoder-only models.

LLaMA-70B and DeepSeek-70B, due to their large size, require three A100 (80GB) GPUs for evaluation, with an estimated runtime of around 7 hours.

## C  Use of AI-Assistant

ChatGPT was used to proofread and improve the text of this paper by correcting grammatical, spelling, and stylistic errors.

# D Fine-grained Results at Paradigm level

| Phenomenon | Paradigm | gemma-3-1b-pt | gemma-3-4b-pt | gemma-3-12b-pt | gemma-3-27b-pt | Llama-3-8B | Llama-3-70B | DeepSeek-R1-Distill-Llama-70B | granite-3.3-2b-base | granite-3.3-8b-base | bloomz-7b1-p3 | Average |
|---|---|---|---|---|---|---|---|---|---|---|---|---|
| **Aspect Agr** | Aspect | 100.00 | 100.00 | 100.00 | 100.00 | 100.00 | 100.00 | 100.00 | 99.25 | 97.51 | 99.25 | 99.60 |
| **Dative-Obj** | Noun | 98.75 | 98.75 | 98.75 | 100.00 | 91.25 | 94.38 | 86.25 | 87.50 | 87.50 | 99.38 | 94.25 |
|  | Pronoun | 98.48 | 97.73 | 97.73 | 99.24 | 93.94 | 98.48 | 90.15 | 91.67 | 90.15 | 93.18 | 95.08 |
| **Ergativity** | Verb.PEF | 95.14 | 95.04 | 98.32 | 98.81 | 98.02 | 99.70 | 98.32 | 91.77 | 97.92 | 97.13 | 97.02 |
|  | DOM | 82.93 | 84.92 | 88.91 | 92.02 | 97.34 | 98.45 | 97.56 | 78.94 | 85.59 | 88.91 | 89.56 |
|  | Verb-Ob Agr | 84.00 | 90.50 | 93.00 | 94.00 | 79.00 | 91.00 | 84.50 | 75.50 | 82.50 | 86.50 | 86.05 |
| **Exp-Subj** | Oblique Pronoun | 79.01 | 84.20 | 94.07 | 84.20 | 98.27 | 95.80 | 96.30 | 77.04 | 91.11 | 67.65 | 86.77 |
| **Honorific** |  | 73.20 | 73.86 | 84.97 | 84.31 | 93.46 | 98.04 | 99.35 | 83.66 | 90.20 | 84.31 | 86.54 |
| **Adj-Noun Agreement** |  | 93.00 | 93.00 | 96.50 | 97.00 | 89.50 | 94.00 | 90.50 | 83.00 | 85.00 | 87.50 | 90.90 |
| **Oblique** | Adjective | 84.31 | 81.37 | 92.16 | 94.12 | 72.55 | 85.29 | 88.24 | 70.59 | 81.37 | 88.24 | 83.82 |
|  | Plural | 88.35 | 83.50 | 89.32 | 90.29 | 85.44 | 95.15 | 88.35 | 89.32 | 92.23 | 77.67 | 87.96 |
|  | Pronoun | 92.00 | 90.00 | 94.50 | 96.30 | 82.70 | 90.40 | 87.70 | 72.30 | 73.50 | 86.60 | 86.60 |
|  | Noun.SG.M | 75.00 | 75.00 | 77.00 | 84.00 | 85.00 | 93.00 | 95.00 | 81.00 | 79.00 | 74.00 | 81.80 |
|  | Verb | 97.06 | 98.04 | 98.04 | 99.02 | 100.00 | 100.00 | 99.02 | 98.04 | 98.04 | 98.04 | 98.53 |
| **Participial Relatives** |  | 79.07 | 81.06 | 85.71 | 85.38 | 90.03 | 92.69 | 88.70 | 80.07 | 82.72 | 81.06 | 84.65 |
| **Subj-Verb Agre** | Gender | 80.90 | 80.90 | 89.14 | 92.51 | 67.04 | 83.15 | 83.52 | 70.04 | 78.28 | 83.15 | 80.86 |
|  | Number | 93.69 | 93.69 | 94.17 | 97.09 | 77.18 | 85.44 | 82.04 | 70.87 | 77.18 | 92.23 | 86.36 |
|  | Person | 88.20 | 92.13 | 94.43 | 95.08 | 85.57 | 92.79 | 93.44 | 53.44 | 61.97 | 78.36 | 83.54 |
| **Order Variation** |  | 82.18 | 86.14 | 91.09 | 91.09 | 90.10 | 94.06 | 90.10 | 85.15 | 93.07 | 85.15 | 88.81 |

Table 7: Fine-grained evaluation results of each syntactic paradigm using *pre-trained models*. The final column reports the average accuracy of each paradigm across all models.

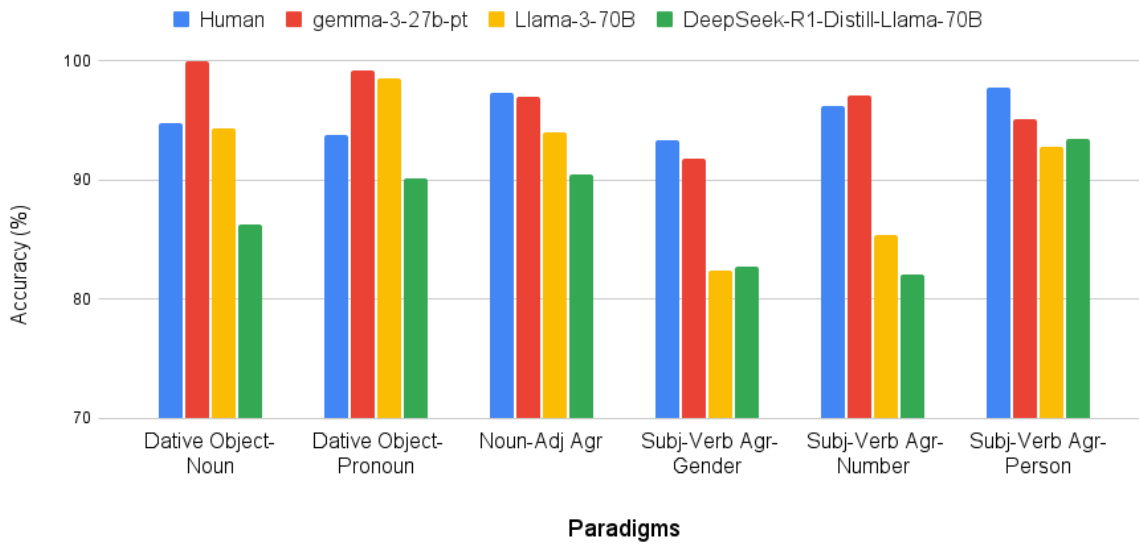

Figure 2: Comparison of model accuracy across linguistic phenomena, showing that the small model outperforms larger models on several linguistic phenomena.

| Phenomenon | Paradigm | gemma-3-1b-it | gemma-3-4b-it | gemma-3-12b-it | gemma-3-27b-it | Alif | granite-3.3-8b-it | granite-3.3-2b-it | Average |
|---|---|---|---|---|---|---|---|---|---|
| **Aspect** | Aspect | 99.00 | 99.75 | 100.00 | 100.00 | 100.00 | 98.75 | 99.00 | 99.50 |
| **Dative-Obj** | Noun | 92.50 | 94.38 | 95.62 | 95.62 | 93.12 | 87.50 | 80.00 | 91.25 |
| | Pronoun | 87.12 | 90.91 | 94.70 | 93.18 | 89.39 | 90.91 | 81.06 | 89.61 |
| **Ergativity** | Verb.PEF | 76.02 | 91.77 | 95.74 | 98.71 | 98.61 | 93.95 | 87.61 | 91.77 |
| | DOM | 56.54 | 81.60 | 86.92 | 90.24 | 98.23 | 84.48 | 74.94 | 81.85 |
| | Verb-Obj Agr | 77.50 | 86.50 | 92.50 | 93.00 | 84.00 | 74.00 | 78.00 | 83.64 |
| **Experiencer-Subj** | Oblique Pronoun | 62.72 | 56.54 | 76.05 | 56.05 | 98.02 | 60.74 | 92.10 | 71.75 |
| **Honorific** | | 55.56 | 74.51 | 79.74 | 84.97 | 98.04 | 94.77 | 88.89 | 82.35 |
| **Adj-Noun Agreement** | | 76.50 | 77.00 | 89.50 | 92.50 | 89.00 | 80.50 | 74.00 | 82.71 |
| **Oblique** | Adjective | 69.61 | 78.43 | 88.24 | 89.22 | 83.33 | 78.43 | 62.75 | 78.57 |
| | Plural | 66.02 | 81.55 | 78.64 | 84.47 | 90.29 | 85.44 | 81.55 | 81.14 |
| | Pronoun | 58.50 | 59.80 | 64.10 | 75.50 | 89.00 | 65.70 | 57.90 | 67.21 |
| | Noun.SG.M | 66.00 | 70.00 | 76.00 | 79.00 | 92.00 | 61.00 | 69.00 | 73.29 |
| | Verb | 90.20 | 94.12 | 92.16 | 97.06 | 99.02 | 99.02 | 99.02 | 95.80 |
| **Participial Relatives** | | 57.48 | 74.42 | 78.74 | 85.38 | 91.69 | 74.09 | 78.41 | 77.17 |
| **Subj-Verb Agreement** | Gender | 57.30 | 71.54 | 84.64 | 91.01 | 80.90 | 77.53 | 69.29 | 76.03 |
| | Number | 81.55 | 90.78 | 94.17 | 95.15 | 77.67 | 69.42 | 63.11 | 81.69 |
| | Person | 72.46 | 84.26 | 91.15 | 91.48 | 90.82 | 66.23 | 50.82 | 78.17 |
| | Order Variation | 63.37 | 68.32 | 78.22 | 83.17 | 97.03 | 91.09 | 76.24 | 79.63 |

Table 8: Fine-grained evaluation results of each syntactic paradigm using ***Instruction-tuned models***. The final column reports the average accuracy of each paradigm across all models.

# E Data Validation Results

Paradigm level human validation results of UrBLiMP are shown in Table 9

| Phenomenon | Paradigm | Accuracy % |
|---|---|---|
| Aspect Agreement | | 100.00 |
| Dative Object | Noun | 94.79 |
| | Pronoun | 93.75 |
| Ergativity | Perfective Verb | 98.75 |
| | DOM | 97.92 |
| | Object Verb Agreement | 96.46 |
| Experiencer Subject | Oblique Pronoun | 93.89 |
| Honorific | Honorific | 94.17 |
| Noun Adjective Agreement | | 97.29 |
| Oblique | Adjective | 92.29 |
| | Plural Noun | 97.71 |
| | Pronoun | 96.88 |
| | Singular Masc Noun | 92.92 |
| | Verb | 97.50 |
| Participial Relatives | | 95.63 |
| Subject Verb Agreement | Gender | 93.33 |
| | Number | 96.18 |
| | Person | 97.71 |
| Word Order | | 98.75 |
| **Average** | | **96.10** |

Table 9: Average human validation accuracy across paradigms. The last row reports the overall average accuracy across all paradigms.

# F UrBLiMP Examples

This appendix contains examples from each of the 19 paradigms in UrBLiMP as shown in Table 10

| Phenomenon | Paradigm | N | Grammatical Sentence | Ungrammatical Sentence |
|---|---|---|---|---|
| Aspect | | 400 | ہاں وہ گورنر سے ملتا رہا تھا۔<br>hā voh gavarnar se milta: raha: tha:<br>Yes he governor from meet-IMPF PROG.PART was<br>'Yes, he had been meeting the governor.' | ہاں وہ گورنر سے ملتا چکا تھا۔<br>hā voh gavarnar se milta: cuka: tha:<br>Yes he governor from meet-IMPF PERF.PART was<br>'Yes, he had already met the governor.' |
| Dative Object | Noun | 160 | پہرےدار نے تمام ماجرا راجا کو سنایا۔<br>pahreda:r ne tama:m ma:jra: ra:ja: ko suna:ya:<br>guard ERG whole incident Raja DAT narrated.PST<br>'The guard narrated the whole incident to Raja.' | پہرےدار نے تمام ماجرا راجا سنایا۔<br>pahreda:r ne tama:m ma:jra: ra:ja: suna:ya:<br>guard ERG whole incident Raja narrated.PST<br>'The guard narrated the whole incident to Raja.' |
| | Pronoun | 131 | آج مجھ کو فرصت ہے<br>a:j mujh ko fursat hai<br>today me.DAT free.time is<br>'Today I have free time.' | آج مجھ فرصت ہے<br>a:j mujh fursat hai<br>today me free.time is<br>'Today I have free time.' |
| Ergativity | Perf.Verb | 1009 | پہرےدار نے تمام ماجرا راجا کو سنایا۔<br>pahreda:r ne tama:m ma:jra: ra:ja: ko suna:ya:<br>guard ERG whole incident Raja DAT narrated.PST<br>'The guard narrated the whole incident to Raja.' | پہرےدار نے تمام ماجرا راجا کو سناتا۔<br>pahreda:r ne tama:m ma:jra: ra:ja: ko suna:ta:<br>guard ERG whole incident Raja DAT narrates.PRES<br>'The guard narrates the whole incident to Raja.' |
| | DOM | 451 | آخر آپا نے منجھلی لڑکی کو آزمایا<br>a:khir a:pa: ne manjhli: laRki: ko a:zma:ya:<br>finally sister ERG middle-aged girl DAT tested.PST<br>'Finally, sister tested the middle-aged girl.' | آخر آپا نے منجھلی لڑکی کو آزمائی<br>a:khir a:pa: ne manjhli: laRki: ko a:zma:'i:<br>finally sister ERG middle-aged girl DAT tests.PRES<br>'Finally, sister tests the middle-aged girl.' |
| | Verb-Obj Agr | 200 | ایک کسان نے چار ایکڑ گندم بوئی<br>ek kisa:n ne ca:r aikR gandum bo'i:<br>one farmer ERG four acre wheat sowed.PST<br>'One farmer sowed four acres of wheat.' | ایک کسان نے چار ایکڑ گندم بویا<br>ek kisa:n ne ca:r aikR gandum boya:<br>one farmer ERG four acre wheat sowed.PST<br>'One farmer sowed four acres of wheat.' |
| Expe-Subj | Obl.Pronoun | 405 | مجھے یہ بلاگ پوسٹ پسند آیا۔<br>mujhe yeh blog post pasand a:ya:<br>to.me this blog post liking came<br>I liked this blog post. | میں یہ بلاگ پوسٹ پسند آیا۔<br>maĩ yeh blog post pasand a:ya:<br>I this blog post liking came<br>I liked this blog post. |
| Honorific | | 153 | بہن جی مجھ سے ناراض تھیں<br>behan ji mujh se nara:z thĩ<br>sister from.me upset was (hon.)<br>Sister was upset with me. | بہن جی مجھ سے ناراض تھی<br>behan ji mujh se nara:z thi<br>sister from.me upset was<br>Sister was upset with me. |
| N-J Agr | | 200 | ہر مجسمہ بیس فٹ اونچا ہے۔<br>har mujasma bi:s fit o:nca: hai<br>every statue twenty feet tall is<br>Every statue is twenty feet tall. | ہر مجسمہ بیس فٹ اونچی ہے۔<br>har mujasma bi:s fit o:nci: hai<br>every statue twenty feet tall.F is<br>Every statue is twenty feet tall. |
| Participial Relatives | | 301 | آسانی سے کھولا گیا دروازہ بند کیا جا سکتا ہے<br>a:sa:ni: se khola: gaya: darwaza: band kiya: ja: sakta: hai<br>easy-ADV from open-PTCP go-PTCP.M.SG door close do-PERF.M.SG go can be<br>'The door that was easily opened can be closed.' | آسانی سے کھولتا گیا دروازہ بند کیا جا سکتا ہے<br>a:sa:ni: se kholta: gaya: darwaza: band kiya: ja: sakta: hai<br>easy-ADV from open-IPFV.M.SG go-PTCP.M.SG door close do-PERF.M.SG go can be<br>'The door that kept opening easily can be closed.' |
| Subj-Verb Agr | Gender | 267 | اور اس کا باپ تو شاید پاگل ہو جاتا<br>aur us ka ba:p to shayad pagal ho jata:<br>and 3SG.M GEN.M father FOCUS perhaps mad become COND.M.SG<br>'And his father might have gone mad.' | اور اس کا باپ تو شاید پاگل ہو جاتی۔<br>aur us ka ba:p to shayad pagal ho jati:<br>and 3SG.M GEN.M father FOCUS perhaps mad become COND.F.SG<br>'And his father (incorrectly) might have gone mad (feminine verb).' |
| | Number | 206 | مٹکا تلاش کیا تو وہ غائب تھا<br>matka: tala:sh kiya: to vo ghaib tha:<br>pot search did then he absent was.SG<br>'When I searched for the pot, it was gone.' | مٹکا تلاش کیا تو وہ غائب تھے<br>matka: tala:sh kiya: to vo ghaib the<br>pot search did then he absent was.PL<br>'When I searched for the pot, they were gone.' (number disagreement) |
| | Person | 305 | میں اسے شوق سے پڑھتا ہوں۔<br>mEn ise shauq se parhta: ho:n<br>I him passion with read.1SG.M be.1SG<br>'I read it with passion.' | میں اسے شوق سے پڑھتا ہو۔<br>mEn ise shauq se parhta: ho<br>I him passion with read.1SG.M be.3SG<br>'I read it with passion.' (person mismatch) |
| Order Variation | | 101 | جو بات ہے وہ بولو۔<br>jo ba:t hai vo bolo<br>what matter is that say.IMP.PL<br>'Say what the matter is.' | جو ہے وہ بولو بات۔<br>jo hai vo bolo ba:t<br>what is that say matter<br>'Say what is that matter.' (incorrect order) |

Table 10: One representative example of grammatical and ungrammatical sentence pairs is shown for each syntactic paradigm. The minimal difference between each pair is underlined. N denotes the total number of sentence pairs included in each paradigm.